# Analysis of Generalized Hebbian Learning Algorithm for Neuromorphic Hardware Using Spinnaker

Shivani Sharma and Darshika G. Perera
Department of Electrical and Computer Engineering, University of Colorado Colorado Springs
Colorado Springs, Colorado, USA
email: ssharma7@uccs.edu, darshika.perera@uccs.edu

*Abstract*—Neuromorphic computing, inspired by biological neural networks, has emerged as a promising approach for solving complex machine learning tasks with greater efficiency and lower power consumption. The integration of biologically plausible learning algorithms, such as the Generalized Hebbian Algorithm (GHA), is key to enhancing the performance of neuromorphic systems. In this paper, we explore the application of GHA in large-scale neuromorphic platforms, specifically SpiNNaker, a hardware designed to simulate large neural networks. Our results demonstrate significant improvements in classification accuracy, showcasing the potential of biologically inspired learning algorithms in advancing the field of neuromorphic computing.

## I. Introduction

Neuromorphic computing, inspired by biological processes, has led to the creation of highly interconnected synthetic neurons and synapses, which are utilized to model neuroscience theories and tackle complex machine learning challenges [1]. The neuromorphic architectures are distinguished by their high connectivity and parallelism, low power requirements, and the integration of memory and processing functions [1]. This combination of biological inspiration and computational power positions neuromorphic computing as a promising frontier for the future of technology [1].

The advantages of these architectures extend beyond their technical specifications, making neuromorphic computing an increasingly compelling area of research give more ways to write this sentence. [33] This system mimics the energy-efficient processing of the human brain, suitable for applications where power consumption is crucial. [33] The highly parallel nature of neuromorphic systems allows for the simultaneous processing of multiple inputs, resulting in faster and more efficient computations. This can adapt to varying inputs and environments, making them robust and versatile across numerous applications [33].

Aligned with these principles, the Generalized Hebbian Algorithm (GHA) further strengthens the potential of neuromorphic computing. By offers numerous advantages as an extension of the classical Hebbian learning rule: GHA effectively identifies principal components of input data, useful for dimensionality reduction and feature extraction [10]. This capability is particularly advantageous for managing large input sets with relatively few outputs, a critical requirement for large-scale neuromorphic systems [35]. Being biologically inspired, aligns with neuromorphic computing principles, offering a natural approach to learning and adaptation [36]. The integration of neuromorphic architectures with GHA showcases how biologically inspired methods drive the development of energy-efficient, scalable, and flexible computing technologies.

SpiNNaker (Spiking Neural Network Architecture) emerges as a specialized hardware platform designed for large-scale neural simulations, offering several ideal features for neuromorphic computing [34]. It supports massive parallel processing, enabling real-time simulation of millions of neurons. This hardware can accommodate large neural networks, making it suitable for implementing and testing algorithms like GHA [34]. SpiNNaker also mimics the energy-efficient processing of the human brain and consumes less power than traditional computing systems, making it ideal platform for Neuromorphic computing [34].

Despite significant progress in deep learning and conventional neural networks, these systems often depend on non-biological learning methods, limiting the full potential of brain-inspired computing [31]. A key challenge is the development of biologically plausible learning algorithms that can be efficiently integrated into hardware platforms like SpiNNaker [32]. Hebbian learning Algorithm (HA)) , a concept introduced by Donald Hebb, provides a biologically inspired method for neural network training [4]. However, traditional Hebbian learning has its drawbacks, particularly in error correction and practical application in complex systems [11]. Addressing these issues, GHA extends the classical Hebbian rule to identify the eigenvectors of the input distribution's autocorrelation matrix [10], making it a promising candidate for neuromorphic applications.

Our objective was to enhance GHA capabilities: We conducted quantitative and functional analysis to assess error rate, average convergence rate, training time, memory usage and classification accuracy for Generalized Hebbian Learning and Hebbian learning models using MNIST dataset and UCI

Machine Learning Repository on neuromorphic hardware, SpiNNaker.

The GHA model demonstrated efficiency, optimization, and effectiveness, improving classification accuracy on the UCI repository with SpiNNaker. By meeting these objectives, this research aims to advance neuromorphic computing, contributing to the development of more efficient and biologically plausible learning algorithms. The study's findings could pave the way for future innovations in neuromorphic systems, enhancing their applicability and performance in various real-world scenarios.

This paper is divided as follows: In Section 2, we discuss Neuromorphic Computing, Hebbian learning Algorithm and Generalized Hebbian learning Algorithm model, in detail. In section 3, we talk about experiments and analysis, including the comparative analysis of GHA model over UCI Machine Learning repository dataset with and without SpiNNaker. For Section 4, we conclude and summarize the entire research done so far and Section 5 discuss about future scope of the Hebbian learning Algorithm.

## II. BACKGROUND

### A. Neuromorphic Computing

The term neuromorphic computing was coined in 1990 by Carver Mead [2]. The neuromorphic architectures are notable for being highly connected and parallel, requiring low-power, and collocating memory and processing [1].

The field of neuromorphic computing encompasses a diverse range of researchers, including those specializing in materials science, neuroscience, electrical engineering, computer engineering, and computer science. Professionals in computer science and engineering focus on developing innovative network models that draw inspiration from both biological systems and machine learning techniques [1]. Their work involves creating new algorithms that enable these models to learn independently and constructing the necessary software to facilitate the practical application of neuromorphic computing systems [1].

The aspiration to design low-power neuromorphic systems has been a significant driving force in this domain, becoming particularly influential approximately a decade into the field's evolution [1].

### B. Hebbian learning Algorithm

Hebbian learning Algorithm (HA) is a learning process which is biologically plausible and ecologically valid. Hebbian learning Algorithm generally works on 'units that fire together, wire together' [12]. The Long-Term Potentiation (LTP) and Long-Term Depression (LTD) helps to learn at neural level [13]. The Long-term potentiation (LTP), a synaptic enhancement, it follows a brief and high-frequency electrical stimulation in the hippocampus and neocortex region of the human brain [14], whereas the Long-term depression (LTD) is synaptic plasticity in which the strength of synaptic transmission between neurons is decreasing gradually. The decrease in synaptic accuracy or efficacy may happen through various mechanisms, for example, a decrease in neurotransmitter receptors on postsynaptic membrane or a lower release of neurotransmitters from the presynaptic neurons. LTP and LTD are widespread phenomena, that express the possibility of excitatory synapse in the mammalian brain [15].

Long-term depression (LTD) was first discovered by Masao Ito in 1982 [16][17]. LTP is induced when the strong input is activated simultaneously with the weak input or following the activation by no more than 20 ms, meanwhile LTD is induced when the temporal order is reversed [18]. This is to show that relative timing is extremely crucial in LTP and LTD induction [19] LTP is also known as Hebbian Plasticity [20]. A typical induction of LTD is usually due to the prolonged low-frequency stimulation of synapses and plays a crucial role in processes such as learning ability, memory loss or memory retention, and in the refinement of neural circuits. It is a complementary process to long-term potentiation (LTP), another form of synaptic plasticity leading to the enhancement of synaptic transmission [21]. Hebbian learning Algorithm is a type of synaptic plasticity that depends on neural activity. It involves the simultaneous activation of presynaptic and postsynaptic neurons, which strengthens their connection [9]. This process enhances the neural response patterns to specific inputs. When these responses are beneficial, their reinforcement can improve accuracy, fluency, and the learning of related perceptions, emotions, thoughts, or actions, contributing to better performance with practice in various tasks. However, if the response is triggered by an undesired input, this reinforcement can have negative effects. The step-by-step algorithm of Hebbian learning Algorithm rule is as follows [7]:

| Algorithm - Hebbian Learning Algorithm | |
|---|---|
| **DATA:** | UCI ML repository Dataset loaded from the TensorFlow library |
| **RESULT:** | Positive or Negative Review based on user inputs |
| **STEP 1:** | Import Initialize all weights and bias to zero, i.e., $w_i=0$ for $i=1$ to $n$, $b = 0$. Here, n is the number of input neurons. |
| **STEP 2:** | For each input training vector and it's respective target output pair s:t, do steps 2–5. |
| **STEP 3:** | Set activation for input units: $x_i = S_i$, $i = 1, …, n$. |
| **STEP 4:** | Set activation for output unit: $y = t$. |
| **STEP 5:** | Adjust the weights and bias: $w_i(new) = w_i(old) + x_i y$ for $i = 1, …, n$, $b(new) = b(old) + y$. If the bias is an input signal that is always 1, the weight change can be written as $w(new) = w(old) + \_w$, where $\_w = xy$. |

A function named Hebbian algorithm that takes three parameters: inputs ( input vector ), target ( output vector ), epochs (number of training iterations) and initializes n ( number of input neuron) as well as set bias and weights vector to zero using the Hebbian Algorithm and produces updated weight as output after training. This function implements a basic Hebbian learning Algorithm, where weights are adjusted based on the correlation between input and output patterns observed during training. Adjustments are made iteratively over multiple epochs to improve the network's ability to predict or classify based on given inputs [7].

### C. Generalized Hebbian learning Algorithm

Generalized Hebbian learning Algorithm is an algorithm to train neural networks to find the eigenvectors of the autocorrelation matrix of the input distribution, given only samples from that distribution [10]. Each output of a trained network represents the response to one eigenvector, and the outputs are ordered by decreasing eigenvalue. A network trained in this way will allow linear reconstruction of the original input with minimal mean-squared error [10]. The step-by-step algorithm of Generalized Hebbian learning Algorithm rule is as follows:

| Algorithm - Generalized Hebbian Learning Algorithm | |
|---|---|
| DATA: | UCI ML repository Dataset loaded from the TensorFlow library |
| RESULT: | Positive or Negative Review based on user inputs |
| STEP 1: | Initialize the weight matrix **C** with small random values. $\mathbf{C}(0) \in R^{M \times N}$<br>Here, M is the number of output neurons, and N is the number of input neurons with M<N. |
| STEP 2: | For each input training vector **x**(t) and its respective target output t, do steps 3–6. |
| STEP 3: | Compute the output vector **y**(t):<br>$\mathbf{y}(t) = \mathbf{C}(t)\,\mathbf{x}(t)$ |
| STEP 4: | Compute the outer products:<br>$\mathbf{y}(t)\mathbf{x}^T(t)$ and $\mathbf{y}(t)\mathbf{y}^T(t)$ |
| STEP 5: | Set all elements above the diagonal of $\mathbf{y}(t)\mathbf{y}^T(t)$ to zero, making it lower triangular: $L^T[\mathbf{y}(t)\mathbf{y}^T(t)]$ |
| STEP 6: | Update the weight matrix **C**(t) using the GHL rule:<br>$\mathbf{C}(t+1) = \mathbf{C}(t) + \eta(\mathbf{y}(t)\mathbf{x}^T(t) - L^T[\mathbf{y}(t)\mathbf{y}^T(t)]\mathbf{C}(t))$ where $\eta$ is the learning rate |

The algorithm starts by initializing the weights matrix C with random values. This matrix will be adapted during training to capture the eigenvectors of the input correlation matrix. Then, it enters a loop for a specified number of training epochs. Within each epoch, it processes each input vector from the dataset. For each input vector x, it calculates the output vector y by taking the dot product of the weight matrix C and the input vector x [10] The core of the algorithm is in updating the weights [10].

Here, Outer product represents the outer product of y and x, and cross product is the cross product of the transpose of C and y. The weights matrix C is then updated based on these products. After completing the specified number of training epochs, the function returns the trained weights matrix C [10]. This algorithm adapts the weights of the neural network based on the input-output relationships, converging to a state where the weights represent the first M eigenvectors of the input correlation matrix, ordered by decreasing eigenvalue [10]. The iterative nature of the training process helps the network learn and capture important patterns in the input data. The convergence is guaranteed by the provided theorem, ensuring that the algorithm finds the sought-after eigenvectors directly from the data without needing to compute the correlation matrix Q in advance [10]

*D.    Neuromorphic Hardware*

Dedicated neuromorphic devices have driven significant research in event-driven models. The emerging neural data communication standard, Address Event Representation (AER), employs packets that use addresses to indicate the source of spikes [29]. This approach efficiently serializes and multiplexes multiple neural signals onto the same lines, simplifying the converters. AER is on the path to becoming a well-defined standard, likely becoming the preferred signaling method for future neural designs. AER signaling is not limited to mixed-signal devices and can serve as a template for neural hardware [29]. The resulting system is a chip with configurable functional blocks, possibly mixed signal, embedded in a connectivity network using AER signaling. This setup utilizes a standard modeling tool chain that also employs an event-driven model, known as the "neurokinetic" architecture [29].

*E.    SpiNNaker*

SpiNNaker, a widely recognized neuromorphic system, is a fully custom digital, massively parallel system [1]. It consists of up to 1,036,800 ARM9 cores and 7 terabytes of RAM distributed across 57,000 nodes, each being a System-in-Package (SiP) with 18 cores and 128 megabytes of off-die Synchronous Dynamic Random Access Memory (SDRAM) [22]. The SpiNNaker engine is designed for massive parallelism with a custom interconnect communication scheme optimized for the spike-based network architecture, handling many small messages efficiently. Each SpiNNaker chip includes instruction and data memory to minimize access time for frequently used data, allowing for the cascading of chips to form larger systems. Although highly flexible in neuron models, synapse models, and learning algorithms, this flexibility incurs energy efficiency costs, consuming 10 nJ per connection as reported by Furber [1].

SpiNNaker is tailored for the "point neuron model [4]," where the dendritic structure of the neuron is ignored, and inputs are applied directly to the soma in the correct temporal order without modeling dendritic tree geometry [22]. The central idea of the SpiNNaker execution model is Address Event Representation (AER) [23][24], where spikes are asynchronous events with information conveyed solely by the spike's timing and the identity of the emitting neuron. AER is implemented by using packet-switched communication along with multicast routing in SpiNNaker, despite introducing some temporal latency, provided it remains well under 1 ms, the error is negligible for modeling biological neural systems [22]. SpiNNaker's communication between neurons utilizes the AER protocol, sending spikes represented by time and identity through a packet-switched communication fabric to connected neurons [25]. This mimics the instantaneous arrival of biological signals in the brain, allowing any neuron to map to any node, forming biological topology irrespective of the communication topology. Efficient mapping has been extensively studied to minimize transmission time between neurons [26]. Communication between ARM cores within a SpiNNaker chip is handled by a Network-on-Chip (NoC), converted to off-chip communication using packet-router modules. Six links are combined using a time-division multiplexer, streaming spikes from the local NoC, then split into six output links. Inter-node communication in SpiNNaker is facilitated via packets generated by cores, transmitted to the local router, and redirected to target cores [25]. If the destination neurons are within the same node, the local router returns the packets to the local cores; if they are in another node, the packets are sent to a neighboring node, requiring an efficient routing technique as each node connects to six other nodes [25]. Multicast routing delivers a single message copy from a source to multiple recipients over a shared communication link [27]. We present the summarized overview of SpiNNaker system in Table I, which is derived from [22].

TABLE I. SUMMARY OVERVIEW OF SpiNNaker SYSTEM

| Component | Details |
|---|---|
| System Type | Massively Parallel Multicore Computing |
| Cores | Up to 1,036,800 ARM9 |
| Memory | 7 TB RAM |
| Nodes | 57,000, each with 18 cores and 128 MB SDRAM |
| Communication Scheme | Custom, optimized for spike-based network architecture |
| Message Handling | Small messages (spikes) |
| Flexibility | Neuron models, synapse models, learning algorithms, and network topology |
| Energy Efficiency | Consumes 10 nJ per connection |
| Neuron Model | Point Neuron Model, inputs applied directly to soma |
| Execution Model | Address Event Representation (AER) |
| AER Implementation | Packet-switched communication and multicast routing |
| Latency | Low temporal latency, negligible error if well under 1 ms |
| Communication Fabric | Electronic, resembling biological signals |
| On-Chip Communication | Managed by Network-on-Chip (NoC) |
| Off-Chip Communication | Converted using packet-router modules |
| Packet Routing | Packets generated by cores, routed to local or neighboring nodes, efficient routing technique needed for inter-node communication |

Packet switching divides the incoming data flow into small packets, which travel through the network similarly to mail, but at much higher speeds [28]. SpiNNaker employs a distributed routing subsystem to direct AER packets across the Communications Network-on-chip. Each chip contains a packet-switching router that efficiently manages and distributes these packets to all connected neurons via the Globally Asynchronous, Locally Synchronous Design and Test interconnect [29]. The routing paths are fully reprogrammable by modifying the routing table, allowing for potential dynamic reconfiguration of the model topology, though this capability remains unexplored [29].

## III. EXPERIMENTAL RESULTS AND ANALYSIS

We performed several experiments to perform comparative analysis including parameters such as average convergence rate, error rate, training time and memory usage for Hebbian learning Algorithm and Generalized Hebbian learning Algorithm. Further, we improved the classification accuracy of Generalized Hebbian learning Algorithm. Detailed experiments and their results are mentioned below. We ran these algorithms on Jupyter notebook and SpiNNaker with Windows 11 (64 bit) specification.

### A. Dataset description

The datasets used for the experiments described below include those from the MNIST and UCI Machine Learning repositories. The MNIST database, which stands for Modified National Institute of Standards and Technology [37], is a substantial collection of handwritten digits frequently employed for training image processing systems [38],[39]. Additionally, it is extensively utilized for training and testing purposes within the machine learning community [40],[41]. This database was formed by combining samples from NIST's original datasets [42].

The UCI Machine Learning Repository is a collection of databases, domain theories, and data generators that are widely used by the machine learning community for empirical analysis of machine learning algorithms. Established in 1987 by UCI PhD student David Aha, the repository has become a primary source of datasets for students, educators, and researchers globally. These datasets provide a valuable resource for testing and comparing the performance of various machine learning algorithms and models. The UCI repository has facilitated countless studies and advancements in the field of machine learning, contributing significantly to the community's development [30]. The Wine dataset from the UCI Machine Learning Repository consists of 178 instances, each with 13 attributes. It is used to classify wines derived from three different cultivars in Italy. The attributes include various chemical properties such as alcohol, malic acid, ash, and color intensity. The Parkinson's Disease dataset includes biomedical voice measurements from 31 people, 23 with Parkinson's disease. The dataset contains 197 instances and 23 attributes that represent different voice measurements, which can be used to distinguish healthy individuals from those with Parkinson's disease. The Heart Disease dataset is used to predict the presence of heart disease in patients. The dataset has 303 instances and 14 attributes, including age, sex, chest pain type, resting blood pressure, serum cholesterol, and maximum heart rate achieved. It helps in understanding the relationship between different factors and heart disease [30]. The Liver Disease dataset, also known as the Indian Liver Patient dataset, contains 345 instances and 7 attributes. It is used to predict liver disease in patients. Attributes include total bilirubin, direct bilirubin, alkaline phosphatase, and alanine aminotransferase. The Breast Cancer dataset is used for the diagnosis of breast cancer. It consists of 286 instances with 9 attributes, which are computed from a digitized image of a fine needle aspirate (FNA) of a breast mass. Attributes describe characteristics of the cell nuclei present in the image, such as radius, texture, perimeter, area, and smoothness [30].

### B. Our Results and Analysis

The GHA and HA Algorithms were first implemented over MNIST (Modified National Institute of Standards and Technology) dataset with and without using SpiNNaker. Both Algorithms were compared with the following factors:

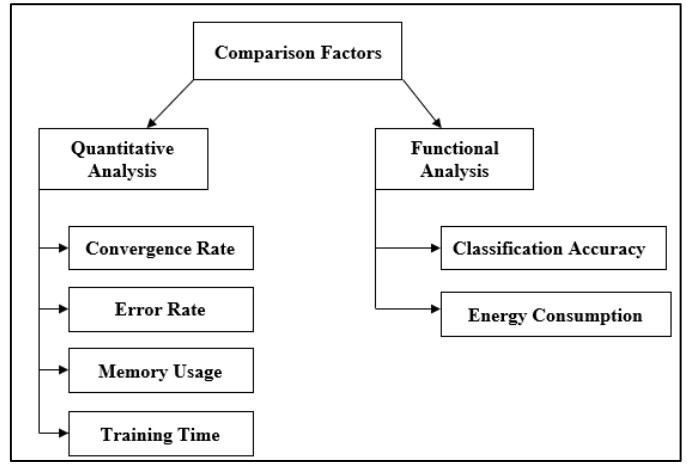

Fig. 1. Comparison Factors for Hebbian learning Algorithm and Generalized Hebbian learning Algorithm.

*1) Quantitative Analysis:* There are three metrics involved : average convergence rate is the measure of quickly algorithm converges to stable solution. Error rate is the measure of how well the network captures and reproduces the input patterns and Computational Complexity includes memory usage and training time to understand the efficiency of the algorithm. The generated results are presented in Table II.

TABLE II. QUANTITATIVE ANALYSIS FOR HEBBIAN LEARNING VS. GENERALIZED HEBBIAN LEARNING ALGORITHMS FOR MNIST DATASET WITH AND WITHOUT SPINNAKER

| MNIST Dataset | Hebbian Learning Algorithm | | Generalized Hebbian Learning Algorithm | |
|---|---|---|---|---|
| | With SpiNNaker | Without SpiNNaker | With SpiNNaker | Without SpiNNaker |
| Error Rate (%) | 52.78 | 86.11 | 16.67 | 19.44 |
| Training Time (sec) | 0.05 | 0.10 | 0.96 | 0.80 |
| Memory Usage(KB) | 0 | 0 | 158484 | 183640 |
| Average Convergence Rate | 0.1556 | 0.1981 | 7.64953 | 4.23490 |

Table II depicts a lower error rate for GHA. When the error rate is lower, it indicates that the model's predictions are closer to the actual values. Therefore, lower error rates generally signify better performance of GHA over HA. Also, a higher average convergence rate of GHA implies that the algorithm is converging towards the optimal or near-optimal solution rapidly. This is desirable as it means the algorithm requires fewer iterations or epochs to learn from the data and produce meaningful results, thus indicating better performance of GHA over HA. Higher memory usage typically suggests that the model is more complex, as it needs to store and process a larger amount of information. This complexity could arise from a higher number of parameters, larger input data, or more sophisticated operations performed during training. Therefore, higher memory usage is often associated with a more complex GHA model than HA model. The higher training time of GHA in above mentioned table suggests that the model is taking longer to learn from the data. This could be due to several reasons, one of which might be the complexity of the model. A more complex model would require more computational resources and time to train because it needs to learn intricate patterns from the data. Therefore, higher training time often indicates a more complex GHA model over HA.

2) *Functional Comparison:* We consider classification accuracy measure for Hebbian learning Algorithm and Generalized Hebbian learning Algorithm to evaluate and understand the findings.

We evaluated the performance of the Generalized Hebbian Learning (GHA) and Hebbian Learning (HA) algorithms using the MNIST dataset. These results are presented in Table III.

TABLE III. CLASSIFICATION ACCURACY OF HEBBIAN LEARNING AND GENERALIZED HEBBIAN LEARNING ALGORITHM WITH SPINNAKER

| MNIST Dataset | Hebbian Learning Algorithm | Generalized Hebbian Learning Algorithm |
|---|---|---|
| Classification Accuracy (%) | 16.67 | 30.56 |

The classification accuracy for the GHA model was notably low, achieving only 30.56%. To enhance the performance, we extended our analysis by applying the GHA model to the UCI-ML datasets. We used the UCI-ML datasets because they are well-established, diverse, and commonly used for benchmarking machine learning algorithms [42].

TABLE IV. CLASSIFICATION ACCURACY AND ENERGY CONSUMPTION OF UCI -ML REPOSITORY ON HEBBIAN LEARNING ALGORITHM WITH TRAINING SET OF 70% AND TESTING SET OF 30% WITH AND WITHOUT SPINNAKER

| UCI Machine Learning Repository | Classification Accuracy (%) | | Energy Consumption (Joule) | |
|---|---|---|---|---|
| | With SpiNNaker | Without SpiNNaker | With SpiNNaker | Without SpiNNaker |
| Wine Dataset | 66.67 | 64.81 | 6200 | 6200 |
| Parkinson's Disease Dataset | 28.81 | 62.71 | 6800 | 6800 |
| Heart Disease Dataset | 62.22 | 55.56 | 10350 | 10350 |
| Liver Disease Dataset | 48.28 | 51.72 | 20250 | 20250 |
| Breast Cancer Dataset | 55.56 | 57.89 | 19900 | 19900 |

TABLE V. CLASSIFICATION ACCURACY AND ENERGY CONSUMPTION OF UCI -ML REPOSITORY ON HEBBIAN LEARNING ALGORITHM WITH TRAINING SET OF 30% AND TESTING SET OF 70% WITH AND WITHOUT SPINNAKER

| UCI Machine Learning Repository | Classification Accuracy (%) | | Energy Consumption (Joule) | |
|---|---|---|---|---|
| | With SpiNNaker | Without SpiNNaker | With SpiNNaker | Without SpiNNaker |
| Wine Dataset | 80.56 | 72.00 | 2650 | 2650 |
| Parkinson's Disease Dataset | 52.55 | 54.74 | 2900 | 2900 |
| Heart Disease Dataset | 43.75 | 46.15 | 4450 | 4450 |
| Liver Disease Dataset | 45.81 | 54.19 | 8650 | 8650 |
| Breast Cancer Dataset | 38.85 | 47.87 | 8500 | 8500 |

TABLE VI. CLASSIFICATION ACCURACY AND ENERGY CONSUMPTION OF UCI -ML REPOSITORY ON HEBBIAN LEARNING ALGORITHM WITH TRAINING SET OF 50% AND TESTING SET OF 50% WITH AND WITHOUT SPINNAKER

| UCI Machine Learning Repository | Classification Accuracy (%) | | Energy Consumption (Joule) | |
|---|---|---|---|---|
| | With SpiNNaker | Without SpiNNaker | With SpiNNaker | Without SpiNNaker |
| Wine Dataset | 41.57 | 34.83 | 4450 | 4450 |
| Parkinson's Disease Dataset | 72.45 | 70.41 | 4850 | 4850 |
| Heart Disease Dataset | 43.62 | 64.43 | 7400 | 7400 |
| Liver Disease Dataset | 49.66 | 50.34 | 14450 | 14450 |
| Breast Cancer Dataset | 52.36 | 42.11 | 14200 | 14200 |

After loading the datasets from the UCI Machine Learning Repository with different datasets, we trained and compiled the GHA (Generalized Hebbian learning Algorithm) model, with and without the use of SpiNNaker. Balancing the training and test split is crucial to ensure that the model is well-trained and accurately evaluated. Since the earlier classification accuracy of GHA came around 30.56%. Different training and testing splits such as 70/30, 50/50, 80/20, and 30/70 for different datasets of UCI Machine Learning Repository on Generalized

Hebbian Algorithm were performed. The results are presented in Tables IV, V, VI, and VII, respectively.

TABLE VII. CLASSIFICATION ACCURACY AND ENERGY CONSUMPTION OF UCI-ML REPOSITORY ON HEBBIAN LEARNING ALGORITHM WITH TRAINING SET OF 80% AND TESTING SET OF 20% WITH AND WITHOUT SPINNAKER

| UCI Machine Learning Repository | Classification Accuracy (%) | | Energy Consumption (Joule) | |
| --- | --- | --- | --- | --- |
| | With SpiNNaker | Without SpiNNaker | With SpiNNaker | Without SpiNNaker |
| Wine Dataset | 69.44 | 72.80 | 7100 | 7100 |
| Parkinson's Disease Dataset | 25.64 | 58.97 | 7800 | 7800 |
| Heart Disease Dataset | 51.67 | 53.33 | 11850 | 11850 |
| Liver Disease Dataset | 51.72 | 43.10 | 23150 | 23150 |
| Breast Cancer Dataset | 38.60 | 39.47 | 22750 | 22750 |

During model training and compilation, we observed significant improvements in classification accuracy. The neuromorphic nature of SpiNNaker allowed for more efficient processing, which translated into better model performance. Highest classification accuracy is 80.56% with 2650 J energy consumption with 30% train set in the wine dataset from UCI machine learning repository with SpiNNaker and Highest classification accuracy is 72.80% with energy consumption of 7100 Joules with 80% train set in the wine dataset from UCI machine learning repository without SpiNNaker. Furthermore, the classification accuracy of the GHA model showed a remarkable improvement. Initially, the GHA model had a classification accuracy of 30.56% with MNIST dataset using SpiNNaker. To improve the classification accuracy of GHA, the main change was the adjustment of the training and test data split. For testing, providing the model with more data to learn from and potentially leading to better performance due to the larger training set. However, with a smaller test set, the evaluation might not be as comprehensive or reliable. Also having the split adjusted to 30% for training and 70% for testing. These improvements highlight the potential of SpiNNaker as a powerful tool for improving the classification accuracy and efficiency of machine learning models. The enhancements in the classification accuracy demonstrate the platform's capability to handle complex datasets and deliver superior results.

## IV. CONCLUSION AND FUTURE WORK

Our analysis reveals that the Generalized Hebbian Algorithm (GHA) outperforms the Hebbian Learning Algorithm (HA) in key performance metrics such as error rate, classification accuracy, and average convergence rate. GHA demonstrates a superior ability to capture underlying data patterns and achieve more accurate predictions while converging to stable solutions more quickly.

However, this enhanced performance comes at the cost of increased training time and memory usage, reflecting the model's complexity. Despite these computational challenges, GHA's strengths in energy consumption and classification accuracy make it well-suited for tasks requiring high performance and rapid convergence, such as image and speech recognition. To fully capitalize on GHA's advantages, it is essential to address its computational demands through efficient resource management. Overall, GHA represents a promising advancement in neuromorphic computing, offering significant improvements in performance that are valuable for complex and large-scale applications.

Future research efforts will focus on further exploring GHA to mitigate its computational overhead while preserving its performance advantages. In this regard, we are planning to investigate FPGA-based architecture for GHA. This is mainly because our previous analyses [43],[44] illustrate that FPGA-based systems are currently the best avenue to support complex (compute/data intensive) applications and algorithms, such as linear Kalmann Filter for two-phase buck converter applications. In addition, our previous work on FPGA-based accelerators, architectures, and techniques for various compute and data-intensive applications, including data analytics/mining [45],[46],[47],[48],[49],[50],[51]; control systems [52],[53],[54],[55]; cybersecurity [56],[57]; machine learning [58],[59],[60]; communications [61]; edge computing [62],[63]; and neuromorphic computing [64]; demonstrated that FPGA-based systems are the best avenue to support and accelerate complex neuromorphic algorithms such as GHA.

Also as future work, we are planning to investigate hardware optimization techniques, such as parallel processing architectures (similar to [60,[65],[66]), partial and dynamic reconfiguration traits (as stated in [67],[68],[69]) and architectures (similar to [57],[70],[71],[72]), HDL code optimization techniques (as stated in [73],[74]), and multi-ported memory architectures (similar to [75],[76],[77],[78]), to further enhance the performance metrics of FPGA-based GHA, while considering the associated tradeoffs.

Overall, our study underscores the significance of Generalized Hebbian learning Algorithm in advancing biologically inspired computing paradigms towards practical implementation and real-world impact.